\documentclass[lettersize,journal,twoside]{IEEEtran}
\usepackage{amsmath,amsfonts}
\usepackage{algorithmic}
\usepackage{array}
\usepackage[caption=false,font=normalsize,labelfont=sf,textfont=sf]{subfig}
\usepackage{textcomp}
\usepackage{stfloats}
\usepackage{url}
\usepackage{verbatim}
\usepackage{graphicx}
\def\BibTeX{{\rm B\kern-.05em{\sc i\kern-.025em b}\kern-.08em
    T\kern-.1667em\lower.7ex\hbox{E}\kern-.125emX}}
\usepackage{balance}

\usepackage[backend=biber,style=ieee,citestyle=numeric-comp,sorting=none,natbib=true,mincitenames=1,maxcitenames=1]{biblatex}
\usepackage{lipsum}
\usepackage[hidelinks]{hyperref}
\usepackage{orcidlink}
\usepackage[capitalise]{cleveref}
\usepackage[nomain,acronym,shortcuts,toc]{glossaries}
\makeglossaries
\newacronym{ai}{AI}{artificial intelligence}
\newacronym{arc}{ARC}{Advanced Research Computing}
\newacronym{asean}{ASEAN}{Association of Southeast Asian Nations}
\newacronym{ann}{ANN}{artificial neural networks}

\newacronym{bit}{BiT}{Big Transfer}

\newacronym{cnn}{CNN}{convolutional neural network}
\newacronym{co2}{CO2}{carbon dioxide}
\newacronym{cli}{CLI}{command-line interface}

\newacronym{dl}{DL}{deep learning}
\newacronym{dfl}{DFL}{Data Fusion Learning}
\newacronym{dvt}{DVT}{Dynamic Vision Transformer}

\newacronym{fot}{FoT}{Forecast Transformer}
\newacronym{fut}{FuT}{Fusion Transformer}
\newacronym{fl}{FL}{federated learning}

\newacronym{gru}{GRU}{gated recurrent unit}

\newacronym{hpc}{HPC}{high-performance computing}
\newacronym{hmtl}{HMTL}{heterogeneous multi-task learning}

\newacronym{iot}{IoT}{Internet of Things}

\newacronym{lstm}{LSTM}{long short-term memory}

\newacronym{mt2v}{MT2V}{multivariate Time2Vec}
\newacronym{ml}{ML}{machine learning}
\newacronym{mae}{MAE}{mean absolute error}
\newacronym{mse}{MSE}{mean squared error}
\newacronym{mape}{MAPE}{mean absolute percentage error}

\newacronym{nlp}{NLP}{natural language processing}
\newacronym{nmt}{NMT}{neural machine translation}
\newacronym{nsf}{NSF}{National Science Foundation}

\newacronym{qos}{QoS}{quality of service}

\newacronym{rnn}{RNN}{recurrent neural network}
\newacronym{resnet}{ResNet}{residual network}
\newacronym{rl}{RL}{reinforcement learning}

\newacronym[longplural={smart and connected communities}]{scc}{S\&CC}{smart and connected community}
\newacronym{scce}{SCCE}{sparse categorical crossentropy}
\newacronym{seq2seq}{Seq2Seq}{sequence-to-sequence}
\newacronym{sgap}{SGAP}{Smart Garden Alley Project}
\newacronym{slurm}{SLURM}{Simple Linux Utility for Resource Management}
\newacronym{sl}{SL}{split learning}
\newacronym{sfl}{SFL}{split federated learning}

\newacronym{t2v}{T2V}{Time2Vec}
\newacronym{tft}{TFT}{Temporal Fusion Transformer}
\newacronym{tst}{TST}{Time Series Transformer}

\newacronym{uav}{UAV}{unmanned aerial vehicle}
\newacronym{unit}{UniT}{Unified Transformer}

\newacronym{vit}{ViT}{Vision Transformer}
\newacronym{vifot}{ViFoT}{Vision Forecast Transformer}
\usepackage[super]{nth}
\usepackage{multirow}
\newcommand{\floor}[1]{\lfloor #1 \rfloor}
\usepackage{longtable} % for 'longtable' environment
\usepackage{pdflscape} % for 'landscape' environment
\usepackage{xcolor}
\usepackage[htt]{hyphenat}
\usepackage{booktabs}
\usepackage{orcidlink}

\AtBeginBibliography{\footnotesize}

\begin{document}

% Define the title within a variable so we can use it later.
\newcommand{\papertitle}{A Transformer Framework for Data Fusion and Multi-Task Learning in Smart Cities}

% Display the title.
\title{\papertitle}

% Display the authors.
\author{%
Alexander C. DeRieux\orcidlink{0000-0003-1606-0668},~\IEEEmembership{Graduate Student Member, IEEE},~%
Walid Saad\orcidlink{0000-0003-2247-2458},~\IEEEmembership{Fellow, IEEE},\\%
Wangda Zuo\orcidlink{0000-0003-2102-5592},~%
Rachmawan Budiarto\orcidlink{0000-0002-5088-0574},~%
Mochamad Donny Koerniawan\orcidlink{0000-0002-1566-9124},~and~%
Dwi Novitasari\orcidlink{0000-0003-0074-0642}
\thanks{This work was funded by the National Science Foundation under Grants CNS-2025377 and CNS-2241361.}
\thanks{A. C. DeRieux and W. Saad are with Wireless@VT, Bradley Department of Electrical and Computer Engineering, Virginia Tech, Blacksburg, VA 24061 USA (email: \{\href{mailto:acd1797@vt.edu}{acd1797}, \href{mailto:walids@vt.edu}{walids}\}@vt.edu).}
\thanks{W. Zuo is with the Department of Architectural Engineering, Penn State University, State College, PA 16801 USA (email: \href{mailto:wangda.zuo@psu.edu}{wangda.zuo@psu.edu}).}
\thanks{R. Budiarto and D. Novitasari are with the Department of Nuclear Engineering and Engineering Physics, Universitas Gadjah Mada, Indonesia (email: \href{mailto:rachmawan@ugm.ac.id}{rachmawan@ugm.ac.id}, \href{mailto:dwi.novitasari@mail.ugm.ac.id}{dwi.novitasari@mail.ugm.ac.id}).}
\thanks{M. D. Koerniawan is with the Department of Architecture, Institut Teknologi Bandung, Indonesia (email: \href{mailto:donny@ar.itb.ac.id}{donny@ar.itb.ac.id}).}
\thanks{We thank Yizhi Yang, John Zhai, Irawan Eko Prabowo, Lily Tambunan, Nissa Aulia Ardiani, Ferdi Mochtar, and Edward Syarif for the various discussions within the context of the Makassar project that helped scope some of the research.}
}

\markboth{IEEE Transactions on Neural Networks and Learning Systems,~Vol.~X, No.~Y, November~2022}{DeRieux \MakeLowercase{\textit{et al.}}: \papertitle}

\maketitle

\begin{abstract}
% Limit 250 words.
%
Rapid global urbanization is a double-edged sword, heralding promises of economical prosperity and public health while also posing unique environmental and humanitarian challenges.
\Acrfullpl{scc} apply data-centric solutions to these problems by integrating \acrfull{ai} and the \acrfull{iot}. This coupling of intelligent technologies also poses interesting system design challenges regarding heterogeneous data fusion and task diversity.
Transformers are of particular interest to address these problems, given their success across diverse fields of \acrfull{nlp}, computer vision, time-series regression, and multi-modal data fusion. This begs the question whether Transformers can be further diversified to leverage fusions of \acrshort{iot} data sources for heterogeneous multi-task learning in \acrshort{scc} trade spaces. 
In this paper, a Transformer-based \acrshort{ai} system for emerging smart cities is proposed.
Designed using a pure encoder backbone, and further customized through interchangeable input embedding and output task heads, the system supports virtually any input data and output task types present \acrshortpl{scc}. This generalizability is demonstrated through learning diverse task sets representative of \acrshort{scc} environments, including multivariate time-series regression, visual plant disease classification, and image-time-series fusion tasks using a combination of Beijing PM2.5 and Plant Village datasets.
Simulation results show that the proposed Transformer-based system can handle various input data types via custom sequence embedding techniques, and are naturally suited to learning a diverse set of tasks. The results also show that multi-task learners increase both memory and computational efficiency while maintaining comparable performance to both single-task variants, and non-Transformer baselines. 
\end{abstract}

\begin{IEEEkeywords}
Artificial Intelligence, Multi-Task Learning, Data Fusion, Transformers, Smart Cities, Internet of Things
\end{IEEEkeywords}

%%%%%%%%%
% I. Introduction

\section{Introduction} \label{se:introduction}

%%%
% Paragraph 1.
% General intro to the topic of the paper, the evolution of the topic, some general references, and the last sentence must say what the challenges are.
%%%
\IEEEPARstart{I}{t} is projected that by 2050 nearly 68\% of the global populace will live in urban areas \cite{WUP2018}. Rapid urbanization in many countries around the world is ushered by promises of economical prosperity and increased population health and wellness. However, these boons pose equal challenges pertaining to environmental quality and food cultivation, both of which are vital to sustaining positive growth. Parallel to this expansion is also an increased emphasis on integrating intelligent technologies to build \acp{scc}. Sparked by recent advancements in \ac{ai} and the \ac{iot}, this convergence of technology and society is the epicenter of \ac{scc} development. 
In particular, the coupling of \ac{ai} with \ac{iot} poses interesting system design challenges pertaining to data collection, and, most importantly, \emph{data fusion}. Specifically, the question of how to design intelligent systems that can leverage a fusion of heterogeneous features arises. This is critical for data-rich locales, such as \ac{scc}, in which many of the environmental features are captured by \ac{iot}, but remain untapped by \ac{ai} to learn correlations.
Hence, next-generation \ac{ai} systems for \ac{scc} environments must be designed to leverage a fusion of heterogeneous data sources to learn multiple tasks concurrently.

Focusing on the perspective of \ac{ai}, there have been many interesting architectural developments to learn from sequence-based information.
Specifically, Transformers \cite{1706.03762} have revolutionized \ac{nlp} and \ac{nmt} trade spaces over competitive \acp{rnn} models (i.e., \ac{lstm}, \ac{gru}, etc.). Moreover, their affinity for sequence-modeling tasks has also been shown to be applicable in fields of time-series regression \cite{1912.09363,2012.07436,2001.08317,2010.02803,2109.12218}, computer vision \cite{2010.11929,2072-4292,21621020}, and multi-modal data fusion \cite{2104.09224,2102.10772}. This applicability across many diverse trade spaces begs the question whether attention-based architectures can be further extended to leverage fusions of \ac{iot} data sources for heterogeneous multi-task learning in \ac{scc} trade spaces.

%%%
% Paragraph 2.
% Literature Review.
%%%

\subsection{Related Works and their Limitations}

There has been a recent influx of literature studying the application of Transformers to time-series forecasting \cite{1912.09363,2001.08317,2012.07436,2010.02803,2109.12218,2205.13504}, computer vision \cite{2010.11929,2072-4292,21621020,2105.15075}, and multi-modal data fusion \cite{2104.09224,2102.10772} trade spaces.
In terms of time-series forecasting, the authors of \cite{1912.09363} propose the so-called \acrfull{tft}, which combines \ac{lstm} technology with attention mechanisms to make quantile forecasts across multiple time horizons.
%
% 2020
In \cite{2001.08317}, the authors propose the first time-series Transformer design consisting of both encoder and decoder blocks to forecast \textit{univariate} temporal data.
% 2020
In \cite{2012.07436}, the authors propose the Informer architecture, which introduces a variation of the traditional attention mechanism, called \textit{ProbSparse} self-attention.
%
% 2020
The authors in \cite{2010.02803} propose an encoder-only Transformer architecture, called the \ac{tst}, for performing both \textit{multivariate} time-series regression and classification tasks.
%
% 2021
In \cite{2109.12218}, the authors develop a further extension of the forecasting Transformer architecture, called Spacetimeformer, which considers both the temporal and spatial relationships between sequence features for \textit{multivariate} time-series regression.
In contrast, the authors in \cite{2205.13504} study the viability of attention mechanisms for the multivariate time-series forecasting task.
Regarding computer vision, the authors of \cite{2010.11929} propose the novel \acrfull{vit} architecture with patch encoding scheme.
%
% 2021
The authors of \cite{2072-4292} successfully apply \ac{vit} architecture to the agriculture trade space, specifically distinguishing crops from weeds using aerial images.
%
% 2021
In \cite{21621020}, the authors also apply the \ac{vit} architecture to agriculture, identifying diseases of casava plants using images of their leafs. 
%
% 2021
In contrast, the authors of \cite{2105.15075} study the validity of vanilla \ac{vit} performance claims.
%
%
%
% 2104.09224
Lastly, regarding data fusion, in \cite{2104.09224}, the authors apply Transformers to fuse images and LiDAR representations of space in autonomous driving tasks.
%
% 2102.10772
Meanwhile, the authors of \cite{2102.10772} propose a novel approach to multi-modal fusion learning, called \ac{unit}, which also doubles as a multi-task learning architecture.

%%%
% Paragraph 3.
% Limitations of Prior Works.
%%%
However, approaches pertaining to time-series forecasting only focus on either univariate regression \cite{1912.09363,2001.08317}, same-feature multivariate regression \cite{2010.02803} (i.e., the features used as input are the same as those predicted at the output), or unique univariate target regression \cite{2012.07436,2109.12218,2205.13504} (i.e., multiple input features and a single unique output feature). 
In terms of computer vision, while the results in \cite{1912.11370} and \cite{2105.15075} demonstrate the overall viability of \ac{vit} variations, and the authors of \cite{2072-4292} and \cite{21621020} prove that \ac{vit} can be applied to agriculture, they only focus on a limited set of images and classes, with the latter only examining diseases in cassava plants.
Regarding data fusion, the limitations of \cite{2104.09224} are two-fold: 1) the reliance on \acp{cnn} for feature extraction, rather than pure Transformer networks, and 2) the similarity between image and LiDAR data sources, both of which represent visual and spatial information. The former being less computationally efficient, and the latter lacking in feature diversity.
Further, the work in \cite{2102.10772} only focuses on image and text inputs, which are not representative of the environmental and temporal features present in \acp{scc}.
Hence, there is a need for novel approaches which not only address these issues, but also consider their broader impact on the global community through application in \ac{scc} environments.

%%%
% Paragraph 4.
% Our Contributions.
%%%

\subsection{Contributions}

The main contribution of this paper is the design of a novel Transformer-based \ac{ai} framework for generalized \ac{iot} data fusion and multi-task learning in \acp{scc}.
Our framework is built upon a backbone of purely Transformer encoder layers, and further customized through interchangeable input embedding and output task heads. This enables system deployments to learn from a variety of heterogeneous input data types from \ac{iot} sources and output task types present \acp{scc} environments. Moreover, the extendable nature of the output heads facilitates learning multiple tasks parallel.
We consider five unique problem sets for our framework, which were chosen to simulate the problem diversity in smart urban environments; these are 1) multivariate time-series forecasting of localized meteorological data, 2) plant health and disease classification using images, a fusion of both 3) multivariate time-series data and 4) images to perform individual forecasting and plant health classification tasks, and 5) a fusion of these same inputs to perform multi-task learning in parallel. 
Deployments of our framework are split across several input type (i.e., single-input, and multi-input) and task type (i.e., single-task, and multi-task) regimes. In the \emph{single-input single-task} regime, a Transformer model is associated with each task and learns it using a single data source (i.e., time-series, images). In the \emph{multi-input single-task} regime, a Transformer model is associated with each task, and learns it using a fusion of image and time-series data sources.
Lastly, in the \emph{multi-input multi-task} regime, a single Transformer model is used for all tasks, which it learns through input data fusion. All of our source code is publicly available on GitHub\footnote{\url{https://github.com/news-vt/makassar-ml}}.

% Paragraph about experiments and results.
We run extensive experiments to evaluate the performance of our framework. 
The results show that the proposed Transformer-based system can indeed handle various input data types via custom sequence embedding techniques. Our results also demonstrate that the Transformer system is well-suited to learning heterogeneous task sets. 
We also compare our Transformer-based models against competitive non-Transformer variants for each task. Specifically, our comparisons show that our proposed framework either maintains or exceeds the performance capabilities of non-Transformer baselines.
In addition, we observe that multi-task learners exhibit both increased memory and computational efficiency while also maintaining comparable performance to both single-task variants and baselines. This affinity demonstrates the flexibility of Transformer networks to learn from a fusion of \acrshort{iot} data sources, their applicability in \acrshort{scc} trade spaces, and their further potential for deployment on edge computing devices.

%%%
% Last paragraph.
%%%
The rest of this paper is organized as follows. \cref{se:problem_formulation} presents the system model. 
In \cref{se:solution}, we propose Transformer architectures for multi-variate time-series regression, visual plant disease classification, single-task data fusion, and heterogeneous multi-task learning tasks.
\cref{se:results} provides simulation results. Finally, conclusions are drawn in \cref{se:conclusion}.
% II. System model (or Problem Formulation)

\section{System Model}\label{se:problem_formulation} % a.k.a. System Model

We consider the design and deployment of an \ac{ai} framework within \ac{scc} environments in which there are networks of \ac{iot} sensors. These sensors collect a wealth of visual, spatial, and ecological data from within the city. This information paints a picture of how the city is jointly effected by these factors, and their relationship can then be used to bolster positive growth and development.

A real-world example of this kind of environment is Makssar City, Indonesia. Makassar is the \nth{5} largest urban center in Indonesia, with a population of 1.7 million \cite{nsf_2025377}, and aims to become a world-class metropolis through combining both \ac{iot} and \ac{ai} to transform many of its 7,520 alleyways into smart and sustainable gardens. 
Supported by the U.S.\ \ac{nsf} and the U.S.\ Department of State, the Smart Garden Alley Project\footnote{\url{https://sites.psu.edu/sbslab/research/city/smart-garden-alleys/}} proposes to develop \emph{smart} garden alleys following a biomimetic philosophy; this represents the garden alleys, \ac{iot} sensor networks, and \ac{ai}-informed government policy as a collection of cells, nerves, and a brain.

We formulate the design of an \ac{ai} framework for \ac{scc} environments, like Makassar City, as problems of both \emph{data fusion} and \emph{multi-task learning}. The variety of data available to \ac{iot} sensor networks necessitates the design of \ac{ai} that can both input diverse feature types, such as visual and ecological, while also learning relationships between them. Likewise, the variety of tasks that can be performed using the \ac{iot} data, such as classification and forecasting, further necessitates \ac{ai} that can not only implement these heterogeneous tasks, but also learn the relationships between them.

\subsection{Heterogeneous Data Fusion}\label{ss:intro-data-fusion}

We now formulate the problem of heterogeneous data fusion in \ac{scc} environments. 
Next-generation \ac{ai} systems for \ac{scc} must learn to leverage a fusion of \ac{iot} data sources to perform necessary tasks. This facilitates learning relationships between loosely-correlated environmental characteristics, and encourages generalization through diverse fusion of these features.
However, the design of such \ac{ai} systems which use these diverse feature sets is non-trivial. In particular, special considerations must be made for both data preprocessing and the overall \ac{ai} architecture. 

With regards to preprocessing, the frequency of each input feature set matters significantly for ingestion into the model. Consider for example time-delimited meteorological data as one feature set, which may have an hourly frequency. If the task necessitates fusion with images, the question arises of how to combine static image frames with this time-delimited data. The key point here is that, while many algorithmic approaches to this fusion are viable, special consideration must be made to formulate a preprocessing algorithm in general. This in turn adds increased layers of complexity to the overall system design.

With regards to the overall \ac{ai} architecture, several considerations must be taken into account when fusing diverse feature sets. Extrapolating from the aforementioned data frequency concern, this also affects the design of model input layers as well as any embedding schemes. In particular, traditional \ac{ai} design assumes inputs to be supplied in discrete increments, and the frequency the data affects the overall input shape. Consider the previous example in which time-delimited data is fused with images. If static images are supplied, then an input layer must be designed to accept just a single image. However, if the preprocessing algorithm instead produces multiple images in a buffer, then the design of the input layer must change to support a sequence of frames. Likewise, knowledge of the input shape will affect the downstream embedding schemes, which could completely change the \ac{ai} architecture. For example, single-image embedding could pool the matrix pixels, whereas multi-image embedding could instead project the sequence of frames. The key point is that the fusion of diverse feature sets is non-trivial, and affects the entire \ac{ai} architecture design.

\subsection{Multi-Task Learning using Attention}\label{ss:intro-multitask-learning-using-attention}

\begin{figure}[t]
	\centering
	\includegraphics[width=\columnwidth]{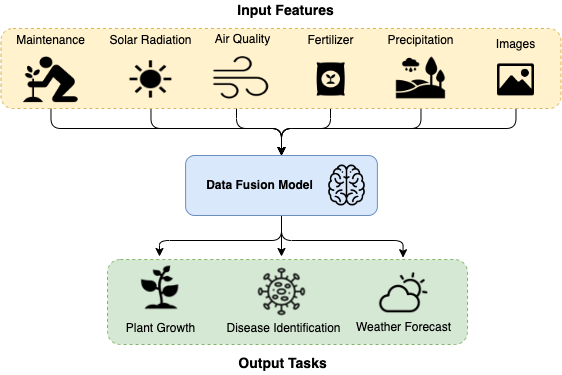}
	\caption{An example of Data Fusion Learning in smart city environments. Input features such as city maintenance, solar radiation, air quality, fertilizer, precipitation, and images can fused to learn plant growth, disease identification, and meteorological forecasting tasks.}
	\label{fig:data_fusion_learning}
\end{figure}

Now, we formulate the problem of multi-task learning of diverse task sets in \ac{scc} environments using attention-based Transformer architectures.
Highly-performant \ac{ai} language models such as BERT \cite{bert}, XLNet \cite{xlnet}, and GPT-3 \cite{gpt3} all use attention mechanisms and Transformers as their baseline architecture. The dominance of Transformers has extended beyond fields of \ac{nlp}, finding applications in computer vision with models such as the \ac{vit} \cite{2010.11929}, and time-series forecasting with models such as the \ac{tft} \cite{1912.09363} and Informer \cite{2012.07436}. However, all of these models are designed with a \emph{singular input data type} in mind (i.e., text, images, time-series, etc.). In the presence of multiple data types, none of the aforementioned architectures alone can learn correlations between the heterogeneous feature sets. 
Consider now our system model of \ac{iot} sensor networks in urban environments, as shown in \cref{fig:data_fusion_learning}. There exists many sensory features from these devices, such as meteorological information, traffic camera video feeds, and citizen mobility from wireless networks that can be leveraged to bolster smart city growth and development. 
In addition to diverse feature sets, each data collection is often also associated with more then one task. Consider for example the fusion of time-series meteorological data and images of plants as discussed in \cref{ss:intro-data-fusion}. Using both inputs, it is possible to perform multiple tasks such as forecasting meteorological events, predicting necessary irrigation changes to promote plant growth, and even identifying the onset of disease. 
Next-generation \ac{ai} systems for deployment in \ac{scc} environments must be able to leverage diverse feature sets such as these to make more tailored decisions. As such, one key goal of this paper is to design and implement an \ac{ai} framework which learns to \ac{scc} growth using a fusion of heterogeneous features.

In light of the Transformer's affinity across diverse trade spaces, we propose to exploit them in the context of smart cities, because they are capable of learning heterogeneous feature correlations for a variety of tasks in parallel fashion.
The Transformer assumes inputs to be of an embedded sequential nature, but is agnostic to task-specific nuances. This generalizability makes the Transformer architecture a prime choice for implementing fusion models, as many data types can be represented using sequence-based embedding schemes. 
Hereinafter, we refer to the process of learning relationships between these diverse feature sets as \emph{data fusion}, and the art of learning multiple tasks as \emph{multi-task learning}. Data fusion correlates features from multiple input sources via timestamp or geographic location and fuses them within the Transformer network. Multi-task learning extends this relation of input features to learning several tasks across various trade spaces in parallel. During training the network optimizes losses for each output task concurrently, allowing a single model to learn from shared experience and thereby boosting generalization performance. 
Several challenges arise in the process of designing Transformer architectures tailored for data fusion and multi-task learning. Feature diversity in both dimension and physical properties are challenges in data fusion. For example, image and meterological datasets have completely unique shapes (e.g., pixels with RGB color channels, multivariate time-series, etc.) and value units (e.g., pixels in range 0-255, hourly temperature in Celsius, etc.). The challenge is designing a fusion technique which takes these properties into account and learns across appropriate sequence dimensions. Likewise, task diversity is a challenge in multi-task learning. The challenge is designing an \ac{ai} architecture such that its core framework supports the diverse dimensionality requirements of the output tasks (e.g., RGB image generation, 24-hour meteorological forecast, etc.) and capability for various multivariate projections (e.g., sinusoidal, exponential, etc.).
Next-generation \ac{ai} systems for \ac{scc} deployments can greatly benefit from both data fusion and multi-task architecture design choices to leverage the wide array of \ac{iot} sensor data available.

In the following section, we address the aforementioned challenges and propose Transformer architectures for data fusion and multi-task learning.
% III & IV. Our Solution (could be one Section III, or two Sections III and IV)

\section{Transformer Networks for Smart Cities}\label{se:solution}

We propose five Transformer-based architectures that are tailored specifically to smart city task sets. In particular, we focus on five specific supervised-learning problems which simulate the tasks encountered in \acp{scc}, like Makassar City. These tasks are divided into two different categories for \emph{single-task} and \emph{multi-task} learning. 

In the single-task category, we first formulate a supervised multivariate time-series regression task. Specifically, unique input features within a historical time window are used to predict a separate set of features along a finite-time horizon. 
Second, we formulate a plant health identification task using images labeled with various healthiness categories (i.e., healthy, diseased, etc.). In this task, a single image is used as input to predict a class label.
Third, we formulate a variation of the regression task that fuses both time-series and image datasets to enhance performance. In this task, a model accepts both a single image and a window of multivariate time-series data as input. The model then uses this information to predict a unique set of output features along a finite-time horizon, similar to the single-input case.
Fourth, we formulate a plant health classification task using a fusion of the aforementioned datasts. Here the same image and time-series window are used as input, whereas now the model predicts the probability distribution over a finite set of class labels.

In the multi-task category, we extend the single-task data fusion problems to perform both heterogeneous tasks simultaneously using a single model. Specifically, we fuse both static images and multivariate time-series windows as input, and use this information to predict both the healthiness class label and a set of output features along a finite-time horizon. This configuration allows a single \ac{ai} model to optimize multiple tasks in parallel, while also benefiting from shared experience.

\subsection{Forecast Transformer for Multivariate Regression} \label{ss:forecasting-transformer}

We propose the \ac{fot} architecture for supervised time-series regression of multi-input and multi-output feature spaces. This model was developed in conjunction with an application to the financial trade space in mind. Specifically, forecasting stock market asset valuations using multiple unique feature sets. 
Naturally, this can be directly applied to our smart city problem, as many tasks involve the computation of arbitrary output values as a function of some input. In the case of urban farming, the input could, for example, be multivariate time-series meteorological data, which is then used to predict output features such as plant irrigation requirements and expected crop yield.
A key point of the \ac{fot} design is that it is agnostic to any specific regression task.
Specifically, a clear advantage of this architecture over alternative competitive models is its portability to multiple trade spaces, given generalized time embedding schemes, and its ability to learn arbitrary multivariate sequence relations.

\begin{figure}[t]
    \centering
    \includegraphics[width=\columnwidth]{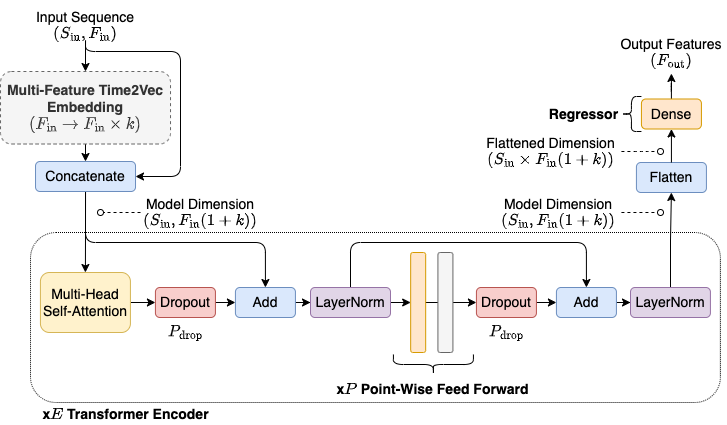}
    \caption{\acrshort{fot} model architecture. Blocks without names are Dense (\textcolor{orange}{orange}), GeLU activation (\textcolor{gray}{gray}), Linear activation (\textcolor{green}{green}), and Dropout (\textcolor{red}{red}).}
    \label{fig:fot_model_design}
\end{figure}

At its core, the \ac{fot} architecture uses a pure-encoder approach to time-series regression. Time-series relations are learned through encoding a multivariate input sequence through cascaded \texttt{TransformerEncoder} layers. These layers employ multi-head self-attention mechanisms \cite{1706.03762} to learn both cross-feature and cross-sequence-index relations within the given input sequence. This is in contrast to the traditional \ac{nlp} Transformer design, which uses both encoder and decoder layers for sequence-to-sequence translation. In addition, the model employs a \ac{t2v} \cite{1907.05321} embedding layer, which learns a vector representation of the input time sequence. The architecture is made dynamic via hyperparameters which control the length of the input sequence, number of input features, number of output features, number of fully-connected sub-layers, embedding dimension, number of cascaded encoder layers, encoder point-wise feed forward dimension, and number of self-attention heads per encoder. The design of the \ac{fot} architecture is shown in \cref{fig:fot_model_design}.

\subsubsection{Multivariate Time2Vec Embedding}\label{ss:time2vec}

Historically, \acp{rnn} have been the tool of choice for sequence-based modeling and regression. The auto-regressive nature of the general \ac{rnn} design allows it to learn inter-element relations using an element's \textit{index}, or more generally \textit{location}, within a given input sequence. In the case of Transformers, entire sequences are processed at once in Attention layers via matrix dot products. This means that the sequence information gathered via auto-regression in RNNs is lost in the presence of Attention-based networks. To overcome this issue for \ac{nlp}-related tasks, the original Transformer architecture adopted a fixed position-encoded embedding scheme, which represents an element's position within a sequence as a sinusoid of varying frequencies \cite{1706.03762}. In working with time-series data, however, this encoding scheme fails to capture the time information of an element relative to the overall dataset (i.e., hourly, daily, or monthly frequencies). We overcame this issue by implementing a variation of the \ac{t2v} embedding scheme proposed in \cite{1907.05321}, which we call \emph{\ac{mt2v} embedding}. This embedding scheme performs two tasks within the network. First, it acts as an embedding layer that represents the input features as vectors in higher-dimensional space. Second, it captures the real time periodicity of each input feature by learning both linear and periodic components and using these as the vector embedding; akin to a learned Fourier Series representation. Mathematically, our \ac{mt2v} embedding scheme extends the definition of \cite{1907.05321} to multivariate input vectors, instead of simple scalar inputs. Given an input feature vector \(\boldsymbol{\tau}\) with \(F_{\textrm{in}}\) feature dimensions, and a total embedding dimension \(k\), the embedding scheme \ac{mt2v} of \(\boldsymbol{\tau}\), denoted as \(\boldsymbol{MT2V}(\boldsymbol{\tau})\), is a 2-dimensional matrix defined as
\begin{align}
    \boldsymbol{MT2V}_{ij}(\boldsymbol{\tau}) = 
    \begin{cases}
        \boldsymbol{\omega}_{ij} \boldsymbol{\tau}_{i} + \boldsymbol{\phi}_{ij}, & \text{if } j = 0, \\
        \mathcal{F}(\boldsymbol{\omega}_{ij} \boldsymbol{\tau}_{i} + \boldsymbol{\phi}_{ij}), & \text{if } 1 \leq j < k,
    \end{cases}
\end{align}
where \(\boldsymbol{\omega}\) is a learned frequency matrix with shape \((F_{\textrm{in}}, k)\), \(\boldsymbol{\phi}\) is a learned phase shift matrix with shape \((F_{\textrm{in}}, k)\), \(\mathcal{F}\) is a periodic function (i.e., \(\sin{}\), \(\cos{}\), etc.), and \(\boldsymbol{MT2V}_{ij}(\boldsymbol{\tau})\) is the \(i\)'th feature and \(j\)'th embedding dimension of the learned embedding matrix with shape \((F_{\textrm{in}}, k)\) for the given input feature vector. In practice, the feature and embedding dimensions are flattened to create a single vector with \(F_{\textrm{in}} \times k\) dimensions. This allows the embedding to be repeated along the sequence dimension \(S_{\textrm{in}}\) to create a final sequence embedding matrix with shape \((S_{\textrm{in}}, F_{\textrm{in}} \times k)\), which is more easily ingested into the Transformer attention layers. The final step of the embedding process, prior to feeding into the Transformer encoder layers, is to concatenate the original batched sequence input with shape \((B, S_{\textrm{in}}, F_{\textrm{in}})\) with the \ac{mt2v} embedding output, resulting in a final encoder input with shape \((B, S_{\textrm{in}}, D_{e})\), where \(D_{e} = F_{\textrm{in}}(1 + k)\) is the inner model dimension of the Transformer encoder.

\subsubsection{Transformer Encoder Layers}\label{ss:fot-encoder}

The core of the network is the Transformer \texttt{Encoder}, which is comprised of cascaded \texttt{TransformerEncoder} layers. The design of these layers is similar to those proposed in \cite{1907.05321}, with some minor changes to be compatible with time-series data. Each layer consists of a two concise sub-layers. The input to the encoder layer is a matrix with shape \((B, S_{\textrm{in}}, D_{e})\), where \(B\) is number of batches, \(S_{\textrm{in}}\) is the input sequence length, and \(D_{e}\) is the inner model dimension (derived from the \ac{mt2v} embedding output shape). In the first sub-layer, the encoder input passes through a multi-head self-attention layer \cite{1907.05321}, which learns, or ``attends'' to, cross-element and cross-feature importance of the input. The output of this attention layer (also a matrix of with shape \((B, S_{\textrm{in}}, D_{e})\)) passes through a dropout regularization layer with probability $P_{\textrm{drop}}$, and then gets added to the original encoder input to form a residual and normalized using a \texttt{LayerNormalization} layer. The output of this normalization passes through the second sub-layer, which consists of a point-wise feed forward network (called \texttt{PointWiseFeedForward}), followed by another sequence of dropout, residual addition, and layer normalization. The point-wise feed forward network acts as a cascaded series of convolutions with a kernel size of 1, which upscales and then downscales the attention output into separate feature dimension $D_{\textrm{ff}}$ respectively. The number of encoders $E$, number of self-attention heads $h$, embedding factor $k$, feed-forward dimension $D_{\textrm{ff}}$, and dropout regularization rate $P_{\textrm{drop}}$ are controlled via hyperparameters.

\subsubsection{Fully-Connected Regressor}\label{ss:fot-regressor}

Up to this point in the network, the model has learned a time vector embedding and cross-feature relations through attention mechanisms in the encoder layers. To actually perform multivariate regression, the encoder output (recall having shape \((B,S_{\textrm{in}},D)\)) must be mapped to the desired output sequence length and number of output features. In this paper we only consider output sequences with length 1 (i.e., 1 day in the future) to simplify the architecture design. However, it should be noted that the design could be extended to support longer sequences in future works. The output feature mapping can therefore be achieved using a single fully-connected layer with \textit{linear} activation as the final layer of the model. This final regressor layer maps the result of the previous fully-connected layers to the desired output feature space (akin to a classification layer in classifier networks, sans softmax activation).

\subsection{Vision Transformer for Plant Health Identification} \label{ss:vision-transformer}

We propose a variation of the well-known \ac{vit} \cite{2010.11929} architecture for computer vision tasks in smart city environments. This facilitates learning image features using attention mechanisms, which maintains a similar architecture structure to that of the aforementioned time-series regression task, and also has been known to exceed performance over competitive \ac{cnn} models.

\begin{figure}[t]
    \centering
    \includegraphics[width=\columnwidth]{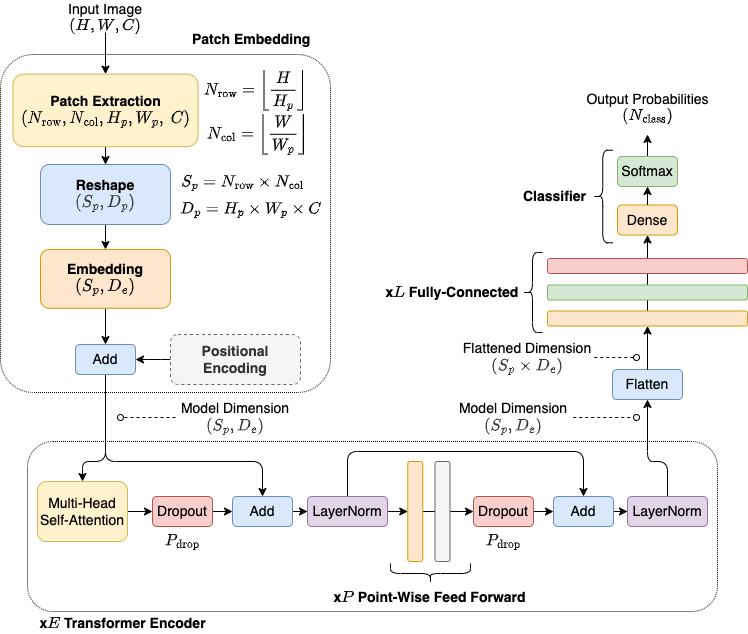}
    \caption{\Acrshort{vit} model architecture. Blocks without names are Dense (\textcolor{orange}{orange}), GeLU activation (\textcolor{gray}{gray}), Linear activation (\textcolor{green}{green}), and Dropout (\textcolor{red}{red}).}
    \label{fig:vit_model_design}
\end{figure}

Similar to the \ac{fot}, the \ac{vit} also leverages the same underlying pure-encoder Transformer architecture for learning image feature correlations. The primary difference between them is in the underlying embedding scheme. Images are single static objects, but the Transformer attention mechanism requires sequence-based input to learn relationships. To make an image compatible as input, the \ac{vit} applies a patch extraction algorithm which converts a static image into a sequence of small square subset frames akin to a checkerboard. These patches are then projected (embedded) into a higher dimensional space matching the input dimension of the cascaded \texttt{TransformerEncoder} layers. The output of the encoders is then passed through a cascaded sequence of fully-connected layers, and subsequently a final dense classification layer for classification tasks.
The \ac{vit} architecture is made dynamic via multiple hyperparameters which control the number of patches, patch shape, number of fully-connected sub-layers, embedding dimension, number of cascaded encoder layers, encoder point-wise feed forward dimension, and the number of self-attention heads per encoder. The design of the \ac{vit} architecture is shown in \cref{fig:vit_model_design}.

\subsubsection{Image Patch Generation}\label{ss:vit-image-patch-generation}

Historically, \acp{cnn} have been used as the standard for image-based \ac{ml} tasks. Convolution-based networks were originally proposed by \citet{10.1109/5.726791} for document character recognition in \citeyear{10.1109/5.726791}, and then popularized in \citeyear{NIPS2012_c399862d} by \citet{NIPS2012_c399862d} for successful application to object classification tasks. 
\ac{cnn} architectures take as input source images with 3 dimensions of $(W, H, C)$, where the former two refer to the image width and height, and the latter the number of color channels (i.e., $C=1$ for grey-scale, $C=3$ for RGB, etc.). Convolutional layers within a \ac{cnn} learn to extract features from source images by performing matrix dot-product operations on small windowed regions commonly referred to as a \textit{kernels}. These kernels are convolved with the input image using various settings for padding and stride offsets which filter the pixels within the region down to a single value. Multiple kernels can be applied in parallel to project the resulting convolved image into higher or lower channel dimensions, thereby varying the number of features that can be extracted from a single image. 

In contrast, the \ac{vit} architecture learns image features by first splitting a source input image into multiple smaller regions called \textit{patches}. The algorithm generates patches similar to a convolution. Patches are extracted using a kernel with shape $(H_{p},W_{p})$ which copies source pixels within its region. This kernel is then convolved with the input image using a stride offset to pad the distance between each region from its nearest neighbors.
The patch extraction operation results in a output matrix with shape $(N_{\textrm{row}},N_{\textrm{col}},H_{p},W_{p},C)$, where $(N_{\textrm{row}},N_{\textrm{col}})$ are the number of rows and columns in the resulting patch grid, $(H_{p},W_{p})$ are the width and height of each patch, and $C$ is the number of channels in the original input image. In practice, the patch matrix for a single image is flattened into shape $(S_{p}, D_{p})$, where $S_{p} = N_{\textrm{row}} \times N_{\textrm{col}}$ serves as the sequence dimension for the Transformer encoder, and $D_{p} = H_{p} \times W_{p} \times C$ serves as the patch projection dimension. Knowing the original patch size and number of color channels allows the user to extract the resulting patches individually for closer inspection. 

\begin{figure}[t]
    \centering
    \includegraphics[width=\columnwidth]{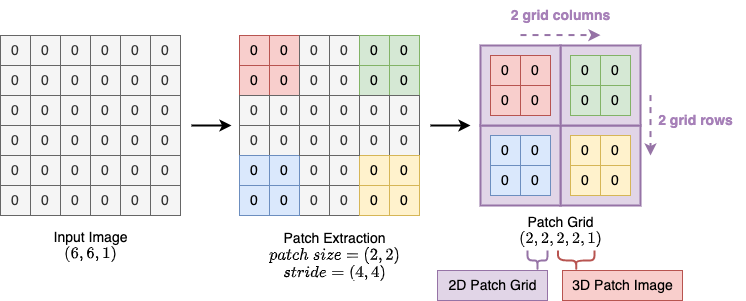}
    \caption{Example of patch extraction algorithm. An input image (left: \textcolor{gray}{gray}) with shape $(6,6,1)$ pixels is split into 4 patches (middle: \textcolor{red}{red}, \textcolor{green}{green}, \textcolor{blue}{blue}, \textcolor{yellow}{yellow}) using patch shape of $(2,2)$ and stride of $(4,4)$. The patches are coalesced into a grid (right) of shape $(2,2,2,2,1)$.}
    \label{fig:patch_example}
\end{figure}

To better understand the patching algorithm a simple example is shown in \cref{fig:patch_example}. In this example, an arbitrary input image with shape $(6,6,1)$ is passed to the extraction algorithm using a patch shape of $(2,2)$ and stride of $(4,4)$. The image is split into 4 unique patches which are organized into a matrix with shape $(2,2,2,2,1)$, where the first 2 dimensions are the patch grid $(2,2)$, and the latter 3 are the patch image $(2,2,1)$.

\begin{figure}[t]
    \centering
    \includegraphics[width=\columnwidth]{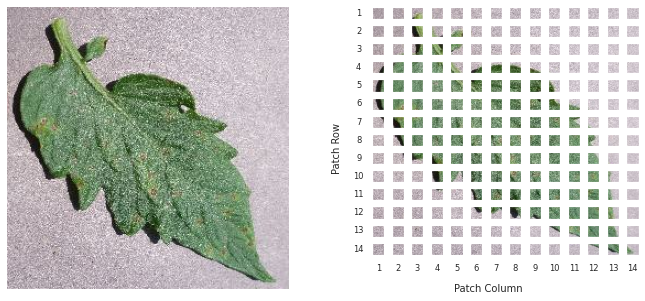}
    \caption{Demonstration of patch extraction algorithm result. Input image (left) with shape $256 \times 256$ pixels is split into 196 patches (right, $14 \times 14$ grid) each with shape $18 \times 18$ pixels.} 
    \label{fig:patch_demo}
\end{figure}

Further, a demonstration of the patch extraction algorithm on a real image is shown in \cref{fig:patch_demo}. Here, an RGB image with shape $(256, 256, 3)$ is used as input. Using a patch size of $(18,18)$, and a stride offset of $(18,18)$, the image is split into 196 unique patches along a $14 \times 14$ grid\footnote{The number patches along each row or column is computed as $N_{\textrm{row}}=\floor{\frac{H}{H_{p}}}$ and $N_{\textrm{col}}=\floor{\frac{W}{W_{p}}}$ respectively}.
Notice that, since the patch size and stride offset are the same, the resulting patches will completely cover the original image with no omitted pixels.

To incorporate the patch extraction algorithm into the \ac{vit} architecture, we designed a static TensorFlow layer called \texttt{Patches}. The advantage of this approach is that the layer can be directly inserted into model build or even preprocessing pipelines to dynamically extract patches from batches of input images. The layer is made dynamic through hyperparameters such as patch size and stride offset which tune the resulting patch dimensions. Note that, by default, the patch size and stride offset are the same so as to have complete coverage of the input image.

\subsubsection{Image Patch Embedding}\label{ss:vit-image-patch-embedding}

Before image patches can be fed into the \texttt{TransformerEncoder} layers they must first be projected into the same inner dimension of the encoder. Recall that output of the patch extraction algorithm for a single image is a matrix of shape $(S_{p}, D_{p})$. However, the Transformer encoder requires the input to have dimension $(S_{p},D_{e})$, where $D_{e}$ is the encoder inner dimension. Therefore, an embedding scheme must be applied which projects the patch dimension $D_{p}$ to the encoder dimension $D_{e}$. In addition, the embedding must also incorporate the sequential information of the patches as the Transformer attention mechanism processes all sequence input concurrently. To accomplish this, we built a custom \texttt{PatchEncoder} layer which does both tasks. First, a fully-connected layer is used to learn the projection $D_{p} \to D_{e}$. Second, a relative positional embedding is added to the projection to encode the sequence information for each patch. The result is a matrix of shape $(S_{p},D_{e})$ which can be fed directly into the Transformer encoder layers.

\subsection{Fusion Transformer for Multi-Task Learning} \label{ss:fusion-transformer}

Up to this point, our proposed model designs used each input data set independently to perform a single task. In particular, \ac{fot} only uses multivariate sequential data for regression, and \ac{vit} only uses images for classification. This compartmentalization of data and tasks results in several \ac{ai} models being required for a single system. Depending on the environment, this separation may be required. In \ac{iot} environments, however, where the data is loosely-correlated (i.e., weather and time-stamped images), it would be more efficient for a single model to input a fusion of heterogeneous input data sets and learn to perform multiple tasks in parallel. 
Hence, we propose the \ac{fut}, which combines the techniques of the aforementioned \ac{fot} and \ac{vit} architectures for both single-task data fusion and multi-task learning. Data fusion is critical in \ac{iot} applications where data collection sources are plentiful, and the data itself is often time-correlated. Examples of such data in smart city environments include weather data from meteorological sensors, and images from video cameras placed on street corners. Using the example of Makassar City's smart garden alleys, the \ac{fut} architecture facilitates building an AI nerve-center for the city that can leverage a fusion of the various \ac{iot} data sources at its disposal to forecast plant health and environmental conditions.

\begin{figure}[t]
    \centering
    \includegraphics[width=\columnwidth]{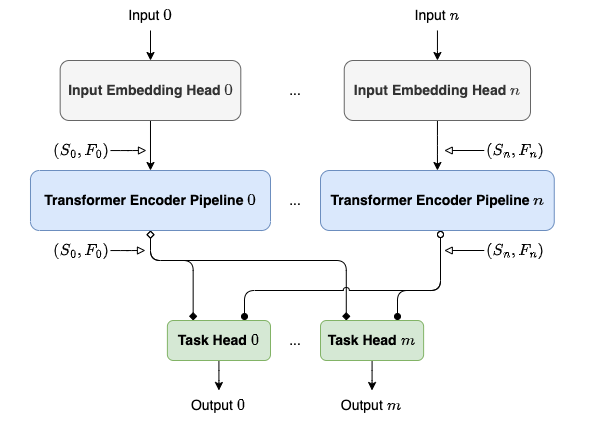}
    \caption{High-level outline of the \acrshort{fut} model architecture for data fusion learning. Each input source has its own embedding head (\textcolor{gray}{gray}) and Transformer encoder pipeline (\textcolor{blue}{blue}). The outputs of each encoder pipeline are then passed through task-specific output head (\textcolor{green}{green}). The number of embedding head and encoder pipeline pairs $n$ is dependent on the amount of input desired, and the number of task heads $m$ is dependent on the application.} 
    \label{fig:fut_model_design_simple}
\end{figure}

The core of the \ac{fut} model is a Y-network where each branch acts as an encoding pipeline that is uniquely associated with each input data source. The output of these pipelines are fed into task-dependent output blocks, referred to as \textit{task heads}, that fuse the learned encodings for all input data sources and applies them to learn a specific task (i.e., classification, regression, etc.). A high-level view of the \ac{fut} architecture is shown in \cref{fig:fut_model_design_simple}. Note that the number of input embedding heads and Transformer encoder pipelines $n$ are tightly coupled, meaning that each input head is directly linked to a specific pipeline. In contrast, the number of task heads $m$ is independent of the input pipelines, and can therefore be chosen relative to the desired tasks.

The interchangeable nature of the task head was specifically designed with transfer learning applications in mind. Transfer learning is a process by which a sub-section of a pre-trained model is grafted into a new architecture for a potentially different application \cite{2104.02144}. This means that the new overall model can benefit from the experience that the pre-trained stub has learned. In \ac{fut}, the encoder pipelines for each input branch can be pre-trained and grafted into a new output head to learn a different task. Further, \ac{fut} supports multi-task learning frameworks with the option for multiple output task heads. This emphasis on both transfer learning and multi-task learning offers many benefits to \acp{scc} as its needs and requirements change over time. For example, initial models could fuse time-series meteorological data and images of plants to forecast irrigation requirements. If the city then desires to add the capability of identifying plant diseases, a classification head can be appended to the pre-trained base model to accomplish both tasks simultaneously.

\subsubsection{Input Embedding Heads}\label{ss:input-embedding-heads}

\begin{figure}[t]
    \centering
    \subfloat[Images]{\includegraphics[width=0.65\columnwidth]{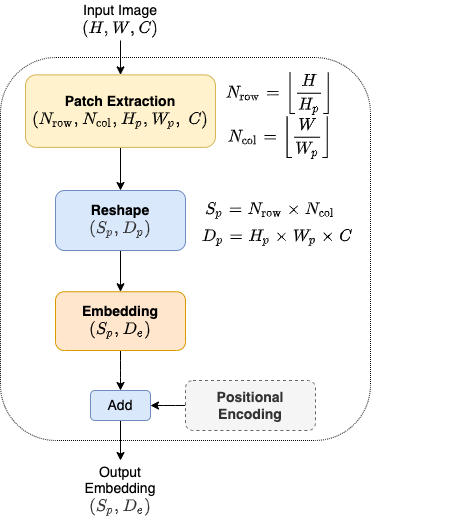}\label{sfig:image_embedding_head}}
    \hfil
    \subfloat[Time-series]{\includegraphics[width=0.35\columnwidth]{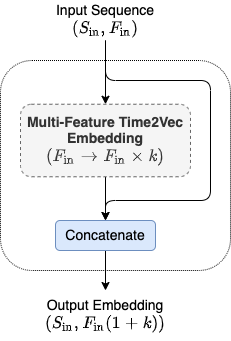}\label{sfig:time_series_embedding_head}}
    \caption{Examples of \acrshort{fut} input embedding heads for (a) images and (b) multivariate time-series datasets.}
    \label{fig:fut_embedding_heads}
\end{figure}

Each branch of the Y-network is dedicated to encoding a specific input data source using cascaded \texttt{TransformerEncoder} layers. The input to each pipeline is a pre-embedded 2-dimensional input sequence. This generalization decouples the input data embedding scheme from the model, thereby making it agnostic to any specific input data sources. It is therefore the job of data-specific input heads to perform embedding and reshaping operations which ultimately conform to the encoding pipeline format. Gleaning from techniques used in \ac{vit} and \ac{fot} Transformers, we extract the embedding sections of these networks and encapsulate them as self-contained embedding layers for image and multivariate time-series data respectively. Examples of these embedding heads are shown in \cref{fig:fut_embedding_heads}.

% \begin{figure}[htb]
% 	\centering
% 	\begin{subfigure}[h]{0.45\textwidth}
%     	\centering
%     	\includegraphics[scale=0.45]{images/image_embedding_head.png}
%     	\caption{Image embedding head}
%     	\label{sfig:image_embedding_head}
%     \end{subfigure}
%     \begin{subfigure}[h]{0.45\textwidth}
%     	\centering
%     	\includegraphics[scale=0.5]{images/time_series_embedding_head.png}
%     	\caption{Time-Series embedding head}
%     	\label{sfig:time_series_embedding_head}
%     \end{subfigure}
% 	\caption{Examples of \acrshort{fut} input embedding heads for images (left) and multivariate time-series datasets (right).}
% 	\label{fig:fut_embedding_heads}
% \end{figure}

The image embedding head in \cref{sfig:image_embedding_head} accepts 3-dimensional images of shape $(H,W,C)$ and extracts patches according to the algorithm discussed in \cref{ss:vit-image-patch-generation}. These patches are then reshaped into a 2-dimensional matrix and embedded according to the technique described in \cref{ss:vit-image-patch-embedding}. Note that here we also add a positional encoding to capture the sequential nature of the image patches in the attention mechanism. This produces a 2-dimensional matrix of shape $(S_{p}, D_{e})$, where $S_{p}$ is the number patches and $D_e$ is the patch embedding dimension. The newly-encoded patches are then passed directly to an associated Transformer encoder pipeline to learn feature relations.

The time-series embedding head in \cref{sfig:time_series_embedding_head} accepts 2-dimensional multivariate sequence data of shape $(S_{\textrm{in}}, F_{\textrm{in}})$, where $S_{\textrm{in}}$ is the sequence length and $F_{\textrm{in}}$ is the number of input features. These features pass through a \ac{mt2v} embedding layer, as discussed in \cref{ss:time2vec}, which represents the time-series information as a Fourier series. This Fourier series representation is then combined with the original input via a concatenation layer along the feature dimension to form a 2-dimensional matrix of shape $(S_{\textrm{in}}, F_{\textrm{in}}(1+k))$, where $k$ is the \ac{mt2v} embedding factor. Similar to the image embedding head, the output of the time-series embedding head is also passed directly to an associated Transformer encoder pipeline to learn feature relations.

A particular point about these input heads is the shape of its associated output. Each produces a 2-dimensional output matrix with shape $(S, F)$, where $S$ is the sequence length and $F$ is the number of embedding features. Notice that the sequence length and feature size are tightly coupled with the input data. This standardization to a 2-dimensional form ensures that all embedding outputs are compatible with Transformer encoder pipelines, which assume 2-dimensional input matrices.

\subsubsection{Data Encoding Pipelines}\label{ss:data-encoding-pipelines}

After the input data passes through each respective embedding head it is fed through a cascaded sequence of \texttt{TransformerEncoder} layers that are unique to the branch. The operation of these layers is similar to those discussed in \cref{ss:fot-encoder}. It is important to note that Transformer encoders maintain the dimensionality of their input data; meaning that the shapes of both input and output matrices are identical. In \ac{fut}, each branch maintains the original shape of its embedding head. This is important because, depending on the embedding scheme, the resulting length and number of features for each sequence could be different. This approach was chosen, however, to make the encoders for each branch unique to the sequences they learn. It is then the job of the following task-specific output head to choose how the sequences are fused.

\subsubsection{Task-Specific Output Heads}\label{ss:task-output-heads}

After the input data passes through its respective Transformer encoder pipelines, the output of each pipeline is then fused together at the final block of the network, referred to as a \textit{task head}. The task head is interchangeable based on the desired \ac{ml} task, and multiple heads can be combined in parallel to build a multi-task learning framework. This interchangeable nature of input heads and ouput heads provides sufficient generalizability to learn the variety of tasks present in smart city environments.

As input, the task head has access to all encoder pipeline outputs at its disposal. The core Y-network learns encoded forms of the original input, and the self-contained task head learns how to fuse these encodings to learn a specific task. Separating the pipelines like this gives the task head full control over how the learned encodings are fused together, which further generalizes the architecture design.

\begin{figure}[t]
    \centering
    \subfloat[Classification]{\includegraphics[width=0.5\columnwidth]{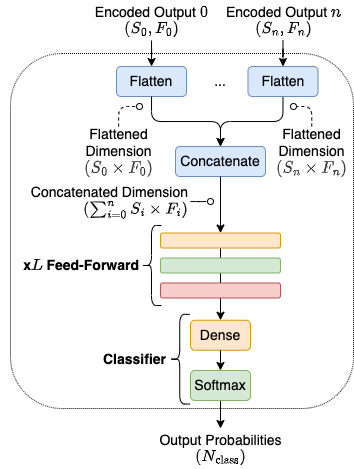}\label{sfig:fut_classification_head}}\hfil
    \subfloat[Regression]{\includegraphics[width=0.5\columnwidth]{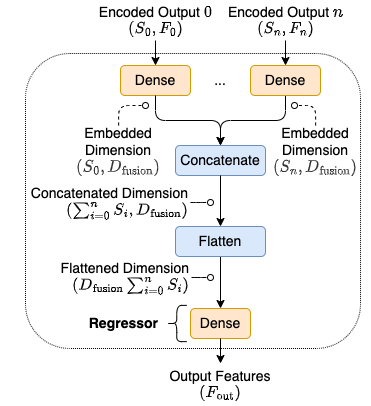}\label{sfig:fut_regression_head}}
    \caption{Examples of \acrshort{fut} output heads for (a) classification and (b) regression tasks. Blocks without names are Dense (\textcolor{orange}{orange}), ReLU activation (\textcolor{green}{green}), and Dropout (\textcolor{red}{red}).}
    \label{fig:fut_output_heads}
\end{figure}

In \cref{fig:fut_output_heads}, we give examples of designs for both classification and regression task heads. Note the similarities between \ac{vit} and \ac{fot} architectures, as discussed in \cref{ss:forecasting-transformer,ss:vision-transformer} respectively. The final task-specific layers of these models have been encapsulated within interchangeable blocks. This allows encoded forms of both input data types to be fused to enhance performance for each task.

% \begin{figure}[htb]
% 	\centering
% 	\begin{subfigure}[h]{0.45\textwidth}
%     	\centering
%     	\includegraphics[scale=0.42]{images/fut_classification_head.png}
%     	\caption{Classification}
%     	\label{sfig:fut_classification_head}
%     \end{subfigure}
%     \begin{subfigure}[h]{0.5\textwidth}
%     	\centering
%     	\includegraphics[scale=0.45]{images/fut_regression_head.png}
%     	\caption{Regression}
%     	\label{sfig:fut_regression_head}
%     \end{subfigure}
% 	\caption{Examples of \acrshort{fut} output heads for classification (left) and regression (right) tasks. Blocks without names are \textit{Dense} (\textcolor{orange}{orange}), ReLU activation (\textcolor{green}{green}), and \textit{Dropout} (\textcolor{red}{red}).}
% 	\label{fig:fut_output_heads}
% \end{figure}

The classification head in \cref{sfig:fut_classification_head} takes the outputs from $n$ encoder pipelines with shapes $(S_0,F_0),\dots,(S_n,F_n)$, flattens them down to 1-dimensional vectors, and concatenates them resulting in a single vector with shape $(S_0 \times F_0 + \dots + S_n \times F_n)$. This vector is then passed through a fully-connected network with ReLU activation and dropout regularization. The final step is a dense classification layer with softmax activation to compute probabilities for each class.

The regression head in \cref{sfig:fut_regression_head} also takes the outputs from the $n$ encoder pipelines with shapes $(S_0,F_0),\dots,(S_n,F_n)$ and projects each independently into a common feature dimension $D_{\textrm{fusion}}$. This results in $n$ sequences with shapes $(S_0,D_{\textrm{fusion}}),\dots,(S_n,D_{\textrm{fusion}})$. This projection is critical as it allows the encoder outputs to be combined along the sequence dimension to form a unified encoded matrix with shape $((S_0 + \dots + S_n, D_{\textrm{fusion}})$. This matrix is flattened and then passed through a fully-connected regression layer with linear activation to compute the output features with dimension $F_{\textrm{out}}$.

There are two important points about these two example task heads. First, notice that both use the data from all encoder pipelines in different ways. Specifically, the classification task head in \cref{sfig:fut_classification_head} flattens each pipeline matrix prior to concatenation, whereas the regression head in \cref{sfig:fut_regression_head} first projects the features of each pipeline matrix to a common dimension and then concatenates these projections along the sequence dimension to form a single unified sequence. The second point is that the outputs of each head are unique and independent of the others, which is subtle yet critical. This means that the model can learn heterogeneous tasks simultaneously using the same input data. In addition, this design also provides the flexibility for the \ac{ai} system to grow in conjunction with the smart city. This means that as the needs of the city evolve over time, so can the \ac{ai} system with the capability for interchangeable output heads. If new tasks are required in the future, custom task heads can be created to generate the necessary output.
% V. Experimental Results

\section{Simulation Results}\label{se:results}

For our simulations, we examine the performance of the \ac{fot}, \ac{vit}, and \ac{fut} architectures presented in \cref{ss:forecasting-transformer,ss:vision-transformer,ss:fusion-transformer} respectively. 
In \cref{se:experiment-fot}, we detail a multivariate time-series regression task to demonstrate the performance of the \ac{fot} model in comparison with a non-Transformer baseline. In \cref{se:experiment-vit}, we detail a plant disease identification task to demonstrate the performance of the \ac{vit} model, and also define a non-Transformer baseline for comparison. In \cref{se:experiment-fusion}, we detail two fusion learning experiments, combining images and time-series data for regression and classification tasks, and demonstrate the performance of the \ac{fut} model with a \emph{single} output head (i.e. single-task mode). In \cref{se:experiment-multitask}, we detail an experiment for multi-task learning using the fusion of images and time-series data to demonstrate the performance of the \ac{fut} model with \emph{multiple} output heads (i.e., multi-task mode). Finally, in \cref{se:experiment-compare}, we compare model performance across all tasks.
All models were trained for 30 epochs and a batch size of 256 on Virginia Tech's \ac{arc} \ac{hpc} cluster.

\subsection{Multivariate Time-Series Forecasting}\label{se:experiment-fot}

In this task, we consider the Beijing PM2.5 \cite{beijing_pm25} dataset, with a history window of 24 hours to predict a horizon of 1 hour. These windows were selected due to their small size (i.e., easier to train) and relative periodicity (i.e., consider what occurred in the past day to predict 1 hour into the future). Specifically, we train the \ac{fot} model to take an input sequence of $S_{\textrm{in}}=24$ consecutive hours with $F_{\textrm{in}}=\{\texttt{TEMP},\texttt{DEWP},\texttt{PRES},\texttt{Iws}\}$ input features per hour, which are \emph{temperature}, \emph{dew point}, \emph{pressure}, and \emph{wind speed} respectively. The \ac{fot} model then predicts a horizon of $S_{\textrm{out}}=1$ hour with $F_{\textrm{out}}=\{\texttt{pm2.5},\texttt{Ir}\}$ output features, which are \emph{air quality}, and \emph{inches of rain} respectively.

To fully examine the performance of our proposed \ac{fot} architecture, we also evaluated a non-Transformer variant as a baseline for comparison. In the case of unique multivariate regression, we considered an \ac{lstm} architecture, which has historically been a top-performer in many regression trade spaces. In terms of parameters, for the \ac{lstm} model we chose 3 cascaded \ac{lstm} layers with 1024, 512, and 256 units each, followed by 2 fully-connected layers with 128, and 64 units each, and 10\% dropout regularization. Note that here we only train a single baseline \ac{lstm} model, instead of hyperparameter tuning. This decision was made intentionally to reduce experiment complexity, directing resources to focus on tuning the proposed models rather than the baselines. We therefore selected this static set of \ac{lstm} network parameters using intuition from previous experience in the financial trade space.

The data was distributed using a train/val/test percentage split of 70/20/10 respectively. This subset size was chosen such that our models would have sufficient historical data to learn periodic trends between seasons. 
The \ac{fot} models were optimized using the Adam optimizer coupled with various learning rate schedules as tuneable parameters to stabilize performance. We minimize \ac{mse} loss and use \ac{mae} as an added metric to indicate performance.

\subsection{Plant Disease Classification}\label{se:experiment-vit}

In this task, we consider images in the Plant Village \cite{plant_village} dataset to predict a probability distribution across 38 healthiness class labels. All images are resized to shape $(72,72,3)$, and scaled to range $[0,1]$. In addition, the training and validation images are augmented with random flips in both the vertical and horizontal axes.

To fully examine the performance of the \ac{vit} architecture, we also evaluated a non-Transformer variant for comparison on the plant disease identification task. Here, we considered the well-known InceptionV3 \cite{1512.00567} architecture, which is based on \acp{cnn}. A ready-to-use implementation of this architecture is provided through the TensorFlow API\footnote{\url{https://www.tensorflow.org/api_docs/python/tf/keras/applications/inception_v3/InceptionV3}}, which we employed to simplify experiment complexity. However, note that for this experiment we trained an InceptionV3 model from scratch; that is, we did not use any pre-trained weight options. This is an important distinction, as TensorFlow provides the option for InceptionV3 variants which have been pre-trained on the ImageNet \cite{10.1109/cvpr.2009.5206848} dataset. In addition, as the InceptionV3 model was provided through TensorFlow, its design is more rigidly defined compared to custom implementations. This prohibits most of the hyperparameter tuning options that we employ with our custom models. Therefore, because of this, we only trained a single InceptionV3 model using the built-in parameter list.

The data was distributed using a train/val/test percentage split of 70/20/10 respectively.
The \ac{vit} models were optimized using the Adam optimizer coupled with a constant learning rate of $10^{-4}$. We chose a constant learning rate here as our preliminary tests indicated that the \ac{vit} model was less sensitive to changes in learning rate, and therefore did not require a decay schedule in contrast to the regression tasks. We minimize the \ac{scce} loss, and we use accuracy as an added metric to indicate performance.

\subsection{Data Fusion Learning}\label{se:experiment-fusion}

In this task, we consider a fusion of the Plant Village and Beijing PM2.5 datasets to ensure fair performance comparison between \ac{fot} and \ac{vit} models. Just like in the classification task, all images are resized to shape $(72,72,3)$, rescaled to range $[0,1]$, and augmented to improve generalization. Likewise for the time-series dataset, we consider an input sequence of $S_{\textrm{in}}=24$ consecutive hours with $F_{\textrm{in}}=\{\texttt{TEMP},\texttt{DEWP},\texttt{PRES},\texttt{Iws}\}$ input features per hour to predict a horizon of $S_{\textrm{out}}=1$ hour with $F_{\textrm{out}}=\{\texttt{pm2.5},\texttt{Ir}\}$ output features.

\subsection{Multi-Task Learning}\label{se:experiment-multitask}

In this task we again consider a fusion of the Plant Village and Beijing PM2.5 datasets to ensure fair performance comparison with \ac{fot}, \ac{vit}, and \ac{fut} models with single output heads. All input data pre-processing for this experiment is equivalent to the single-head fusion experiments discussed in \cref{se:experiment-fusion}.

Because the multi-task \ac{fut} architecture has multiple output heads, each head is given its own loss and metrics to optimize. The regression head uses \ac{mse} loss with \ac{mae} metric, and the classification head uses \ac{scce} loss with accuracy metric. These losses are aggregated using weighted mean into a single loss value for each train/val/test category to optimize the overall model. Model performance can therefore be examined either for each individual head (using its associated loss), or for the entire model using the aggregated loss.

\subsection{Performance Comparison}\label{se:experiment-compare}

\begin{figure}[t]
    \centering
    \subfloat[Regression]{\includegraphics[width=\columnwidth]{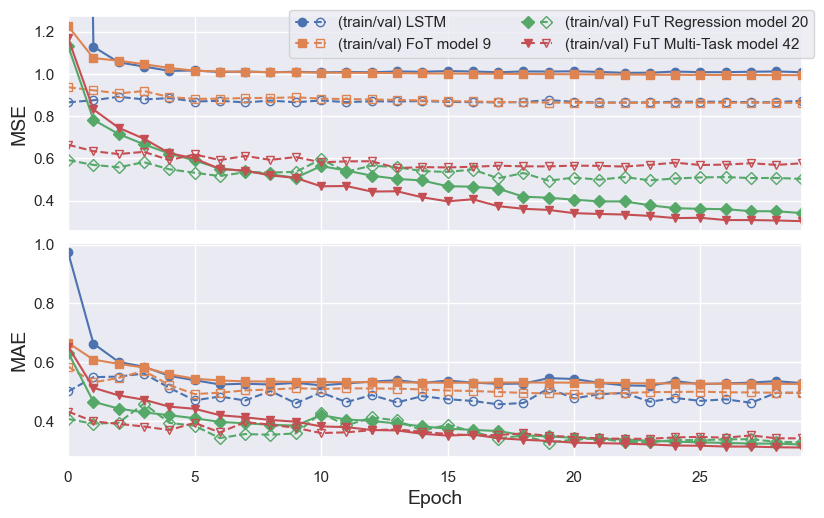}\label{sfig:perf_regression}}
    \\
    \subfloat[Classification]{\includegraphics[width=\columnwidth]{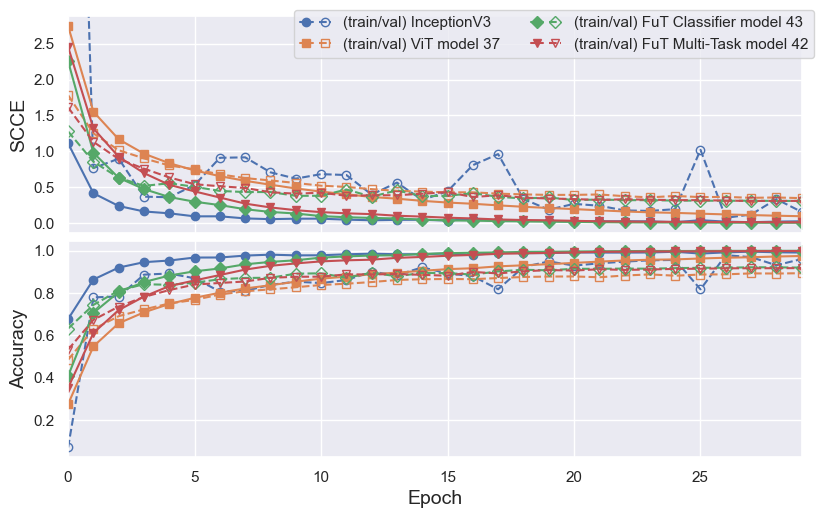}\label{sfig:perf_classifier}}
    \caption{Training and validation performance of best models on (a) classification, and (b) regression tasks.}
    \label{fig:perf}
\end{figure}

\begin{figure}[t]
    \centering
    \subfloat[MAE]{\includegraphics[width=0.5\columnwidth]{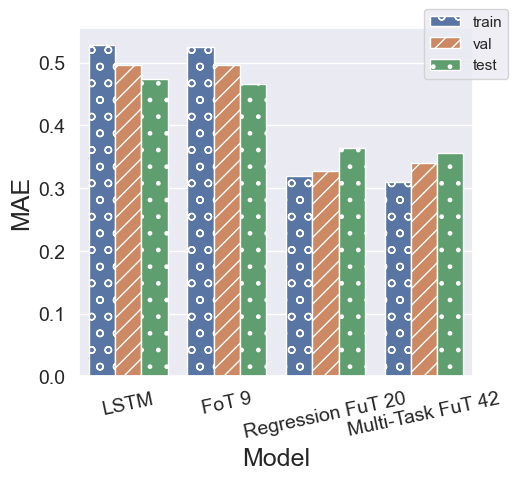}\label{sfig:model_compare_mae}}
    \hfil
    \subfloat[MSE]{\includegraphics[width=0.5\columnwidth]{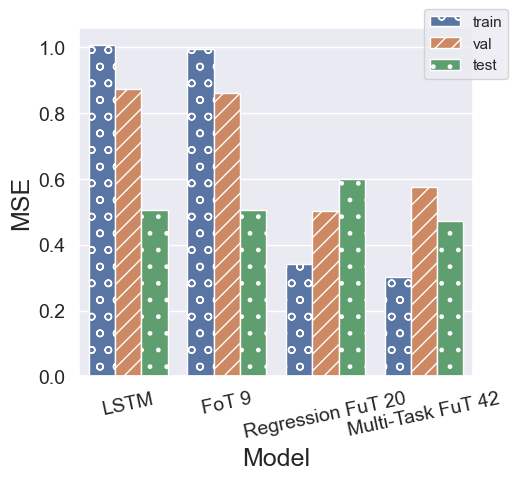}\label{sfig:model_compare_mse}}
    \\
    \subfloat[Accuracy]{\includegraphics[width=0.5\columnwidth]{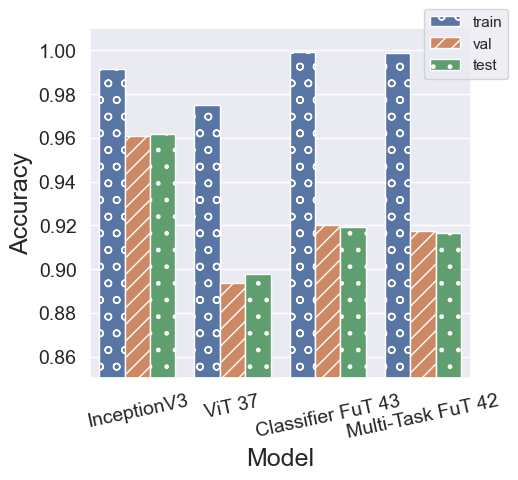}\label{sfig:model_compare_accuracy}}
    \hfil
    \subfloat[SCCE]{\includegraphics[width=0.5\columnwidth]{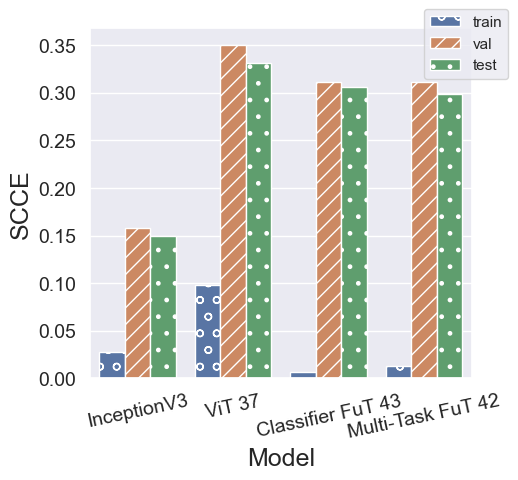}\label{sfig:model_compare_scce}}
    \caption{Comparison between best model metrics for regression (a) \ac{mae} and (b) \ac{mse}, and classification (c) Accuracy and (d) \ac{scce}.}
    \label{fig:model_compare}
\end{figure}

For each of the tasks discussed in \cref{se:experiment-fot,se:experiment-vit,se:experiment-fusion,se:experiment-multitask} we train 48 hyperparameter variations of our \ac{fot}, \ac{vit}, single-task \ac{fut}, and multi-task \ac{fut} models. The best configuration for each model is identified as having the lowest validation loss metric across all variants (i.e., \ac{mse} for regression, and \ac{scce} for classification). We omit the complete training results across all models for brevity, instead concisely listing the best models and their hyperparameters in \cref{tab:top_hyper}. 
Specifically, we found \textbf{\ac{fot} model 9}, \textbf{\ac{vit} model 37}, \textbf{regression \ac{fut} model 20}, \textbf{classifier \ac{fut} model 43}, and \textbf{multi-task \ac{fut} model 42} as the best.
The train/val/test performance for these models across all epochs are shown in \cref{fig:perf}. Likewise, their train/val/test metrics are shown visually in \cref{fig:model_compare}, and tabularly in \cref{tab:model_perf_comparison_regression,tab:model_perf_comparison_classification}. 
Further, while raw performance is important, computational complexity should also be taken into account, especially in \ac{iot} environments where energy and compute resources are at a premium. To finalize the comparison between models we therefore also consider their size (in number of parameters). The total sizes of these best models are shown in \cref{tab:model_size_comparison}. 

\begin{table}[hbt]
    \centering
    \small
    \caption{Hyperparameters of the top-performing models for each task. Unlisted values indicate the associated parameter is not used for the model.}
    \resizebox{\columnwidth}{!}{%
    \begin{tabular}{r *{9}{c}}
        \toprule
        % Model & Parameters \\
        Model 
            & $P_{\textrm{drop}}$ % \texttt{dropout} 
            & $k$ % Sequence embedding dim for FoT and regression FuT
            & $D_e$ % \texttt{embed\_dim} 
            & $D_{\textrm{ff}}$ % \texttt{ff\_dim} 
            & $E$ % \texttt{n\_encoders} 
            & $h$ % \texttt{n\_heads}
            % & \texttt{fc\_units}
            & $(H_p, W_p)$ % \texttt{patch\_size}
            & $D_{\textrm{fusion}}$ % \texttt{fusion\_embed\_dim}
            % & $D_{\textrm{image}}$ % \texttt{image\_embed\_dim}
            % & $k$ % $D_{\textrm{seq}}$ % \texttt{seq\_embed\_dim}
            \\
        \midrule
        % LSTM & \\
        FoT 9 
            & 0.1 % \texttt{dropout=0.1}
            & 5 % \texttt{embed\_dim=5}
            & % D_e
            & 512 % \texttt{ff\_dim=512}
            & 6 % \texttt{n\_encoders=6}
            & 8 % \texttt{n\_heads=8}
            \\[0.7em]
        % \clinelr{1-5}
        % InceptionV3 & \\
        ViT 37 
            & 0.3 % \texttt{dropout=0.3}
            & % k
            & 32 % \texttt{embed\_dim=32}
            & 256 % \texttt{ff\_dim=256}
            & 6 % \texttt{n\_encoders=6}
            & 8 % \texttt{n\_heads=8}
            % & [] % \texttt{fc\_units=[]}
            & (6,6) % \texttt{patch\_size=6}
            \\[0.7em]
        Regression FuT 20
            & 0.3 % \texttt{dropout=0.3}
            & 5 % sequence embed_dim
            & 32 % image embed dim
            & 256 % \texttt{ff\_dim=256}
            & 3 % \texttt{n\_encoders=3}
            & 8 % \texttt{n\_heads=8}
            % & % fc_units
            & (6,6) % \texttt{patch\_size=6}
            & 8 % \texttt{fusion\_embed\_dim=8}
            % & 32 % \texttt{image\_embed\_dim=32}
            % & 5 % \texttt{seq\_embed\_dim=5}
            \\[0.7em]
        Classifier FuT 43
            & 0.3 % \texttt{dropout=0.3}
            & 10 % seq embed dim
            & 32 % image embed_dim
            & 256 % \texttt{ff\_dim=256}
            & 6 % \texttt{n\_encoders=6}
            & 8 % \texttt{n\_heads=8}
            % & % fc_units
            & (6,6) % \texttt{patch\_size=6}
            & % fusion_embed_dim
            % & 32 % \texttt{image\_embed\_dim=32}
            % & 10 % \texttt{seq\_embed\_dim=10}
            \\[0.7em]
        Multi-Task FuT 42 
            & 0.3 % \texttt{dropout=0.3}
            & 5 % seq embed dim
            & 32 % image embed dim
            & 256 % \texttt{ff\_dim=256}
            & 6 % \texttt{n\_encoders=6}
            & 8 % \texttt{n\_heads=8}
            % & % fc_units
            & (6,6) % \texttt{patch\_size=6}
            & 16 % \texttt{fusion\_embed\_dim=16}
            % & 32 % \texttt{image\_embed\_dim=32}
            % & 5 % \texttt{seq\_embed\_dim=5}
            \\
        \bottomrule
    \end{tabular}
    }
    \label{tab:top_hyper}
\end{table}

\begin{table}[hbt]
    \centering
    \captionsetup{position=top} % This is required for tabular children to properly link with the parent table environment.
    \caption{Performance of best trained models on (a) regression, and (b) classification tasks. Best metrics in each column are highlighted in \textbf{bold}.}
    \subfloat[Regression\label{tab:model_perf_comparison_regression}]{%
    \resizebox{\columnwidth}{!}{%
    \begin{tabular}{rrrrrrr}
        \toprule
        % model & mse & val\_mse & test\_mse & mae & val\_mae & test\_mae \\
        \multirow[c]{2}{*}{Model} & \multicolumn{3}{c}{MSE} & \multicolumn{3}{c}{MAE} \\
        \cmidrule(lr){2-4} \cmidrule(lr){5-7}
        & \emph{train} & \emph{val} & \emph{test} & \emph{train} & \emph{val} & \emph{test} \\
        \midrule
        LSTM & 1.0078 & 0.8728 & 0.5044 & 0.5284 & 0.4957 & 0.4745 \\
        FoT 9 & 0.9939 & 0.8630 & 0.5057 & 0.5258 & 0.4956 & 0.4668 \\
        Regression FuT 20 & 0.3414 & \textbf{0.5031} & 0.5999 & 0.3194 & \textbf{0.3270} & 0.3632 \\
        Multi-Task FuT 42 & \textbf{0.3031} & 0.5767 & \textbf{0.4734} & \textbf{0.3092} & 0.3405 & \textbf{0.3554} \\
        \bottomrule
    \end{tabular}
    }}\\
    \subfloat[Classification\label{tab:model_perf_comparison_classification}]{%
    \resizebox{\columnwidth}{!}{%
    \begin{tabular}{rrrrrrr}
        \toprule
        % model & scce & val\_scce & test\_scce & accuracy & val\_accuracy & test\_accuracy \\
        \multirow[c]{2}{*}{Model} & \multicolumn{3}{c}{SCCE} & \multicolumn{3}{c}{Accuracy} \\
        \cmidrule(lr){2-4} \cmidrule(lr){5-7}
        & \emph{train} & \emph{val} & \emph{test} & \emph{train} & \emph{val} & \emph{test} \\
        \midrule
        InceptionV3 & 0.0276 & \textbf{0.1581} & \textbf{0.1490} & 0.9912 & \textbf{0.9606} & \textbf{0.9615} \\
        ViT 37 & 0.0977 & 0.3499 & 0.3307 & 0.9750 & 0.8936 & 0.8978 \\
        Classifier FuT 43 & \textbf{0.0064} & 0.3112 & 0.3057 & \textbf{0.9994} & 0.9202 & 0.9192 \\
        Multi-Task FuT 42 & 0.0135 & 0.3108 & 0.2985 & 0.9987 & 0.9176 & 0.9166 \\
        \bottomrule
    \end{tabular}
    }}
\end{table}

\begin{table}[hbt]
    \centering
    \small
    \caption{Size comparison of best models per input and task types via total number of parameters. Here we also show the size of combined single-task models for comparison with multi-task variants.}
    \label{tab:model_size_comparison}
    \resizebox{\columnwidth}{!}{%
    \begin{tabular}{lllll}
        \toprule
        \multirow{2}{*}{Input Type} & \multirow{2}{*}{Task Type} & \multirow{2}{*}{Model} & \multicolumn{2}{c}{Parameters} \\
        \cmidrule(lr){4-5}
        & & & \textit{Individual} & \textit{Combined} \\
        \midrule
        \multirow{4}{*}{Single-input} 
            & \multirow{4}{*}{Single-task}
                & \ac{lstm} & 8,191,298 & \multirow{2}{*}{30,071,944} \\
            &   & InceptionV3 & 21,880,646 \\
            \cmidrule(lr){3-5}
            &   & \ac{fot} 9 & 167,330 & \multirow{2}{*}{476,712} \\
            &   & \ac{vit} 37 & 309,382 \\
            \cmidrule(lr){1-5}\morecmidrules\cmidrule(lr){1-5}
        \multirow{3}{*}{Multi-input} 
            & \multirow{2}{*}{Single-task} 
                & Regression \ac{fut} 20 & 120,042 & \multirow{2}{*}{651,928} \\
            &   & Classifier \ac{fut} 43 & 531,886 \\
            \cmidrule(lr){3-5}
            & Multi-task & \ac{fut} 42 & \textbf{428,488} & - \\
        \bottomrule
    \end{tabular}
    }
\end{table}

Consider the performance of the regression models in \cref{tab:model_perf_comparison_regression}, \cref{sfig:perf_regression}, and \cref{sfig:model_compare_mae,sfig:model_compare_mse}. 
First, we see that our proposed \ac{fot} model 9 exhibits slightly better training, validation, and testing performance compared to the baseline \ac{lstm}. Specifically, \ac{fut} model 9 achieves train/val/test \ac{mse} $0.9939/0.8630/0.5057$ and \ac{mae} $0.5258/0.4956/0.4668$, whereas the \ac{lstm} achieves \ac{mse} $1.0078/0.8728/0.5044$ and \ac{mae} $0.5284/0.4957/0.4745$.
Second, we see that our proposed single-task \ac{fut} regression model 20 outperforms the previous models in all metrics, while also exhibiting the highest overall validation scores, with \ac{mse} $0.3414/0.5031/0.5999$ and \ac{mae} $0.3194/0.3270/0.3632$.
Finally, we see that our proposed multi-task \ac{fut} model 42 outperforms all other models in terms of training and test performance, with train/val/test \ac{mse} $0.3031/0.5767/0.4734$ and \ac{mae} $0.3092/0.3405/0.3554$. In addition, the lack of plateau in its training and validation performance in \cref{sfig:perf_regression} indicates that increased training could further improve the model.

% All CLASSIFICATION performance.
Next, consider the performance of the classification models in \cref{tab:model_perf_comparison_classification}, \cref{sfig:perf_classifier}, and \cref{sfig:model_compare_accuracy,sfig:model_compare_scce}. Here we see much more variation between training, validation, and testing \ac{scce} loss and accuracy metrics.
First, the InceptionV3 model exhibits the best validation and test scores, with \ac{scce} $0.0276/0.1581/0.1490$ and accuracy $0.9912/0.9606/0.9615$. However, it's convergence in \cref{sfig:perf_classifier} is much more unstable starting around epoch 5.
Second, we see that \ac{vit} model 37 learns the training set rather well, but performs the worst in terms of validation and test metrics, with \ac{scce} $0.0977/0.3499/0.3307$ and accuracy $0.9750/0.8936/0.8978$.
Lastly, we see our proposed single-task \ac{fut} classifier model 43 (\ac{scce} $0.0064/0.3112/0.3057$, accuracy $0.9994/0.9202/0.9192$) and multi-task \ac{fut} model 42 (\ac{scce} $0.0135/0.3108/0.2985$, accuracy $0.9987/0.9176/0.9166$) exhibit very similar performance. In addition, the convergence of all our proposed classifier models exhibits a plateau by epoch 30. This indicates that further training on the image dataset would likely not improve classification performance.

Next, consider the relationship between model size and overall performance for single-task learners with and without data fusion.
Here we see that the \ac{fut} regression model 20 has 47,288 \textit{fewer} parameters compared to \ac{fot} model 9, and 8,071,256 \textit{fewer} parameters than the baseline \ac{lstm}, while exhibiting significantly better performance.
In contrast, the \ac{fut} classification model 43 has 222,504 \textit{more} parameters than \ac{vit} model 37 and has similar performance. This indicates that, while not guaranteed, data fusion can enhance performance, but can also come at the cost of increased complexity. However, comparing with the baseline InceptionV3, we see a dramatic reduction of 21,571,264 \textit{fewer} parameters. As discussed in \cref{se:experiment-fusion}, the baseline InceptionV3 indeed has better performance than \ac{vit}, however, the boost is marginal when also comparing the memory and computational complexities introduced by such a large increase in model size.

Finally, consider the overall performance of multi-task learners with combined single-task variants. To perform multiple tasks using single fusion models, we must combine \ac{fut} regression model 20 and classification model 43. However, doing so requires a total of $120,042 + 531,886 = 651,928$ parameters. If we then compare the results of multi-task \ac{fut} model 42 we see that equivalent performance can be achieved with only 428,488 parameters (a reduction of 223,440). 
If we were to combine the simpler single-input single-task \ac{fot} and \ac{vit} models, totaling $309,382 + 167,330 = 476,712$ parameters, multi-task \ac{fut} model 42 is still smaller (reducing the size by 48,224).
Further, multi-task \ac{fut} model 42 is significantly smaller than the two baseline \ac{lstm} and InceptionV3 models combined, totaling $8,191,298 + 21,880,646 = 30,071,944$ parameters.
This reduction in model size is significant for two reasons. First, the smaller size requires less computational resources for inference, thereby making deployment more viable on \ac{iot} devices. Second, it also reduces the complexity of having two separate models, which further reduces memory overhead, and also reduces deployment costs, as one multi-task \ac{iot} device can do the job of two single-task variants.
These boons in performance, efficiency, and cost therefore establish the multi-task data fusion architecture as the prime candidate for deployment in smart city environments.
% VI. Conclusion

\section{Conclusion}\label{se:conclusion}

In this paper, we have provided several \ac{ai} designs based on pure-encoder Transformer architectures for use in \ac{scc} environments.
We showed that pure-encoder Transformers serve as a viable baseline for single-task learning models, and demonstrated their capability on independent supervised multivariate time-series regression and computer vision classification tasks.
We enhanced our proposed single-task \ac{ai} system design by leveraging the fusion of the aforementioned heterogeneous datasets as input, and showed that this fusion can improve learning over single-input variants.
We further extended the fusion architecture with custom output task heads to facilitate heterogeneous multi-task learning. The flexibility offered by these custom heads allows our proposed architecture to support virtually any task, and any combination of tasks desired.
We showed through extensive experimentation that our proposed multi-input multi-output enhancement either maintains or exceeds the performance capabilities of both single-task variants and non-Transformer baselines, while also reducing its memory footprint and computational overhead. 
Hence, we have demonstrated that our proposed next-generation multi-task Transformer-based \ac{ai} system is viable for deployment in smart city environments. It can successfully learn to perform unique multivariate regression of time-series meteorological features, while concurrently identifying the onset of fungal diseases through visual feature extraction. Moreover, it performs these heterogeneous tasks using a fusion of both static images and multivariate time-series data. Indeed, this flexibility to support diverse feature sets from heterogeneous \ac{iot} data sources, while also learning a variety of tasks in parallel, shows that our system can bolster sustainable urban development practices, and, in particular, foster the growth of \acp{scc}.
%%%%%%%%%

%%%
% Bibliography
%%%
\printbibliography[heading=bibintoc]

%%%
% Biographies
%%%

\vspace{-0.5cm}

\begin{IEEEbiography}[{\includegraphics[width=1in,height=1.25in,clip,keepaspectratio]{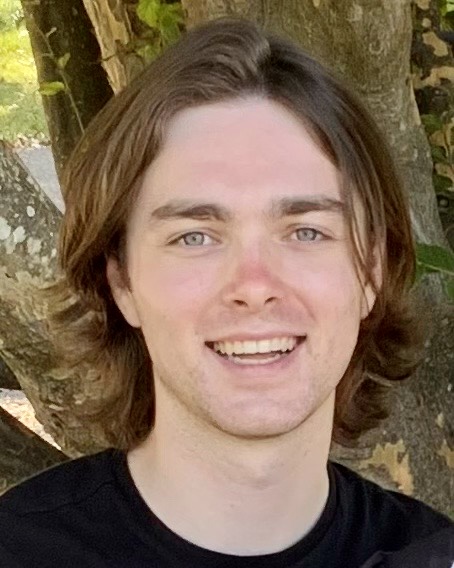}}]{Alexander C. DeRieux} (Graduate Student Member, IEEE) received the B.S.\@ degree in electrical engineering with honors and the B.S.\@ degree in computer science with honors from Virginia Tech, Blacksburg, VA, USA, in 2016, and the M.S.\@ degree in electrical engineering in 2022. He is currently a Ph.D.\@ student in electrical engineering and Bradley Fellow at the Network sciEnce, Wireless, and Security (NEWS) Laboratory in the Bradley Department of Electrical and Computer Engineering at Virginia Tech. His research interests include machine learning, quantum computing, quantum machine learning, quantum communications, and wireless networks.
\end{IEEEbiography}

% \vfill
\vspace{-0.5cm}

\begin{IEEEbiography}[{\includegraphics[width=1in,height=1.25in,clip,keepaspectratio]{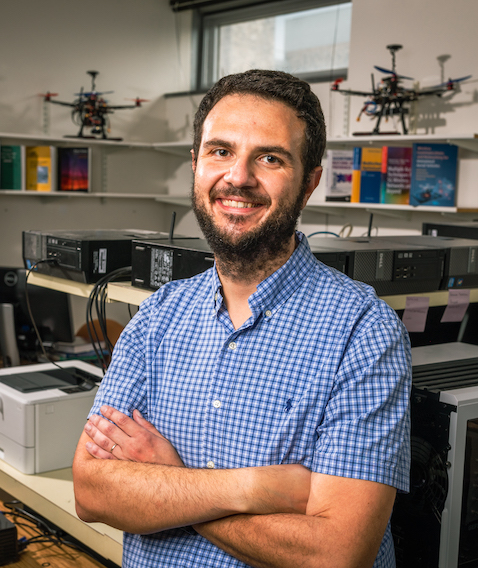}}]{Walid Saad} (Fellow, IEEE) (S’07, M’10, SM’15, F’19) received his Ph.D degree from the University of Oslo in 2010. He is currently a Professor at the Department of Electrical and Computer Engineering at Virginia Tech, where he leads the Network sciEnce, Wireless, and Security (NEWS) laboratory. He is also the Next-G Wireless research leader at Virginia Tech's Innovation Campus. His research interests include wireless networks (5G/6G/beyond), machine learning, game theory, security, unmanned aerial vehicles, semantic communications, cyber-physical systems, and network science. Dr. Saad is a Fellow of the IEEE. He was the author/co-author of eleven conference best paper awards. He is the recipient of the 2015 and 2022 Fred W. Ellersick Prize from the IEEE Communications Society, of the 2017 IEEE ComSoc Best Young Professional in Academia award, of the 2018 IEEE ComSoc Radio Communications Committee Early Achievement Award, and of the 2019 IEEE ComSoc Communication Theory Technical Committee. He was also a co-author of the 2019 and 2021 IEEE Communications Society Young Author Best Paper. He is the Editor-in-Chief for the IEEE Transactions on Machine Learning in  Communications and Networking.
\end{IEEEbiography}

\vspace{-0.5cm}

\begin{IEEEbiography}[{\includegraphics[width=1in,height=1.25in,clip,keepaspectratio]{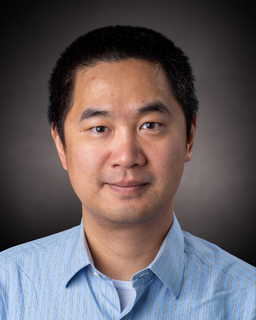}}]{Wangda Zuo} is a Professor in Architectural Engineering and Mechanical Engineering, as well as Associate Director for Research at Global Building Network at Pennsylvania State University. Dr. Zuo also holds a joint appointment at the National Renewable Energy Laboratory (NREL). He is currently an Associate Editor of Journal of Solar Energy Engineering, and Fellow of International Building Performance Simulation Association. He is a major contributor to multiple open-source building and community energy modeling tools, including  Modelica Buildings library and URBANopt.
\end{IEEEbiography}

\vspace{-0.5cm}

\begin{IEEEbiography}[{\includegraphics[width=1in,height=1.25in,clip,keepaspectratio]{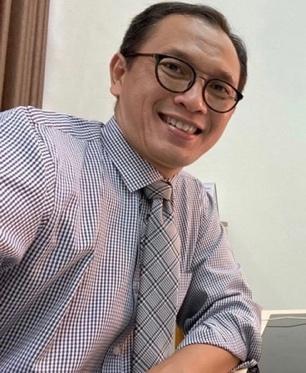}}]{Rachmawan Budiarto} is a faculty member in Department of Nuclear Engineering and Engineering Physics, Director of Centre for Development of Sustainable Region (CDSR) in Centre for Energy Studies as well as a senior researcher at Center for Economic Democracy Studies, Universitas Gadjah Mada. He is a Greenship Professional of Green Building Council Indonesia. He is a member of International Committee of Indonesia Accreditation Board for Engineering Education (IABEE). He was/is involved in at least 81 sustainable energy related projects and 20 other projects, which some of them were/are funded by 17 international institutions.
\end{IEEEbiography}

% \vspace{-0.5cm}
\vfill

\begin{IEEEbiography}[{\includegraphics[width=1in,height=1.25in,clip,keepaspectratio]{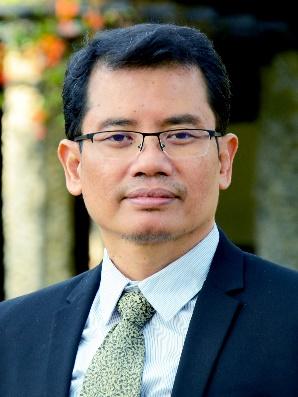}}]{Dr Eng. Mochamad Donny Koerniawan, ST, MT}, lecturer and researcher in Building Technology Research Group within The School of Architecture Planning and Policy Development. Institut Teknologi Bandung, vice president of International Building Simulation Association, Indonesia Chapter; Green Associate and Member of Green Building Council Indonesia (GBCI); member of Association Green Building Expert Indonesia; Member of Indonesia Architect Institute. Highly experienced in green building research and development. Development of Sustainable Region (CDSR) in Centre for Energy Studies.
\end{IEEEbiography}

% \vfill
\vspace{-0.5cm}

\begin{IEEEbiography}[{\includegraphics[width=1in,height=1.25in,clip,keepaspectratio]{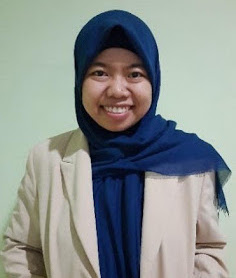}}]{Dwi Novitasari} is a Ph.D. student at the Department of Electrical Engineering and Information Technology, Universitas Gadjah Mada. She is also a researcher at the Centre for Energy Studies and the Centre for Development of Sustainable Region (CDSR), Universitas Gadjah Mada. She was/is involved in national and international projects related to renewable energy, energy planning, energy transition, and sustainable energy. Currently, her research is about climate, land use, energy, and water nexus in power generation planning in Indonesia. She also received the Canada-ASEAN Exchange Scholarship at the School of Sustainable Energy Engineering, Simon Fraser University.
\end{IEEEbiography}

\vfill

\end{document}